\documentclass{article}

\usepackage{arxiv}

\usepackage[utf8]{inputenc} 
\usepackage[T1]{fontenc}    
\usepackage{hyperref}       
\usepackage{url}            
\usepackage{booktabs}       
\usepackage{amsfonts}       
\usepackage{nicefrac}       
\usepackage{microtype}      
\usepackage{lipsum}
\usepackage{graphicx}
\graphicspath{ {./images/} }

\title{People Counting System for Retail
Analytics using Edge AI}

\author{
 Karthik Reddy Kanjula \\
  School of Coumputing and Information Technology\\
  REVA University\\
  Bangalore, KA 560064 \\
  \texttt{karthikreddykanjula@gmail.com} \\
   \And
 Vishnu Vardhan Reddy \\
  School of Coumputing and Information Technology\\
  REVA University\\
  Bangalore, KA 560064 \\
  \texttt{kvvrvishnu007@gmail.com} \\
  \And
Jnanesh K P \\
  School of Coumputing and Information Technology\\
  REVA University\\
  Bangalore, KA 560064 \\
  \texttt{officialjnan@gmail.com} \\
  \AND
  Jeffy S Abraham\\
  School of Coumputing and Information Technology\\
  REVA University\\
  Bangalore, KA 560064 \\
  \texttt{jeffysabraham52@gmail.com} \\
  \And
  Tanuja K \\
  School of Coumputing and Information Technology\\
  REVA University\\
  Bangalore, KA 560064 \\
  \texttt{k.tanuja@reva.edu.in} \\
}

\begin{document}
\maketitle
\begin{abstract}
Developments in IoT applications are playing an important role in our day-to-day life, starting from business predictions to self driving cars. One of the area, most influenced by the field of AI and IoT is retail analytics. In Retail Analytics, Conversion Rates - a metric which is most often used by retail stores to measure how many people have visited the store and how many purchases has happened. This retail conversion rate assess the marketing operations, increasing stock, store outlet and running promotions ..etc. Our project intends to build a cost-effective people counting system with AI at Edge, where it calculates Conversion rates using total number of people counted by the system and  number of transactions for the day, which helps in providing analytical insights for retail store optimization with a very minimum hardware requirements. 
\end{abstract}


\section{Introduction}
Edge computing has been proposed to adapt to the immense measures of information being created, which pushes cloud administrations from the organization center to the organization edges that are in nearer nearness to IoT gadgets and information sources. Indeed, the marriage of edge and AI has offered ascend to another exploration territory, to be specific "edge knowledge" or "edge AI". Rather than altogether depending on the cloud, edge knowledge capitalizes on the far-reaching edge assets to acquire AI understanding. Eminently, edge knowledge has accumulated a lot of consideration from both the business and the scholarly community. For instance, Gartner observed the publicity cycle has fused edge knowledge as an arising innovation that will arrive at a level of profitability in the accompanying 5 to 10 years. Significant ventures, including Google, Microsoft, Intel and IBM, have advanced pilot undertakings to exhibit the benefits of edge processing in clearing the last mile of AI.
\paragraph{}
People counters are basically any kind of gadget that tallies the quantity of individuals strolling all through a specific determined area. While people counters started out as a simple, manual clicker in the hands of a doorman, they have quickly evolved to sensor technologies i.e., [Pressure sensitive mats, Infrared beam counters and thermal counters]. The greater part of the major offline retailers are currently investing resources into cutting edge people counters to level the online-offline battleground and access a similar degree of top to bottom information as internet business brands to comprehend and enhance store transformation and deals.According to Capillary technologies, starting from 2018 to coming future, the people counting systems will get upgraded to Computer Vision and AI-Powered People Counters which will be the most cost effective and requires very minimum hardware. Highly accurate Computer vision models process in cloud to perform inference. As the model inference is needed when a network is not available, this is where “AI at Edge” comes to play. Edge applications are used for real-time decision making and avoid sending data to cloud to perform inference.
\paragraph{}
Intel’s Distribution for OpenVINO Toolkit: OpenVINO [Open Visual Inferencing and Neural Network Optimization] An open-source library useful for edge deployment due to its performance maximization's and pre-trained models. This toolkit helps in optimizing neural network inferencing across a variety of Intel hardware and helps optimizing speed and decrease in model size for inference at Edge. We can train a model in any Framework in cloud and give it Model Optimizer in OpenVINO toolkit which has the ability to converts the models from multiple Frameworks to Intermediate Representation [IR] for Inference Engine which is the main component to perform inference at Edge.

\section{Related work}
\label{sec:headings}
\cite{shi2019} We predict that edge computing will continue to develop rapidly until 2020. After2020, edge computing will step into a steady development period. \cite{zhang} Driving by flourishing both AI and IoT, there is an astringent need to push the AI frontier from the cloud to the network edge. To fulfill this trend, edge computing has been widely recognized as a promising solution to support computation-intensive AI applications in resource-constrained environments. The nexus between edge computing and AI gives birth to the novel paradigm of EI. Recent research in people counting can be divided into techniques using neural-based crowd estimation \cite{tesei}, \cite{leung}, \cite{cho} and methods which are based on blob detection and blob tracking \cite{zabih}, \cite{davis}, \cite{wixson}. By the bottom-up methodology, we partition research endeavors in Edge Computing into Topology, Content, and Service and acquaint a few models on how with stimulate edge with intelligence. By top-down decay, we partition the exploration endeavors in AI anxious into Model Adaptation, Framework Design, and Processor Acceleration and present some current examination results. At long last, we present the cutting edge and amazing difficulties in a few hotly debated issues for both AI for edge and AI on edge \cite{albert}. Identifies critical components of edge intelligence: edge caching, edge training, edge inference. Compares, discuss and analyze the adopted techniques, objectives, performance, advantages and drawbacks \cite{sasu}. Discusses the various technologies used for counting people and points out the best method (blob detection) and explains the process and accuracy rate \cite{dhaief}. The prepared model can be stacked into an AI chip to play out the undertaking of identifying human heads. The test precision can arrive at 98.7\% and the bouncing boxes recognized by the prepared model is exact enough for a constant recognition framework. All in all, when we use complete preparing information, the Mipy assessment board can get the adequate identification precision to play out the useful applications \cite{shen}. People-counting which utilized a blend of recognition and tracking strategies is proposed. The framework utilized a camera for acquiring top-down film of strolling individuals close to the entrance of a room. The detection cycle is finished utilizing a Deep Learning design, a blend of MobileNetv2 and SSD, which accomplished cutting edge brings about object detection \cite{duong}.

\section{Proposed Architecture}
In our proposed system we implement a low resource people counting system which has state of the art accuracy. In this work we use intel's  OpenVINO architecture which deploys the AI model on an inference engine which can give instant results in real time. The model can generate output on a basic intel's CPU with no GPU's or TPU's required.
\begin{figure}[!ht] 
    \centering
    \includegraphics[width=15cm]{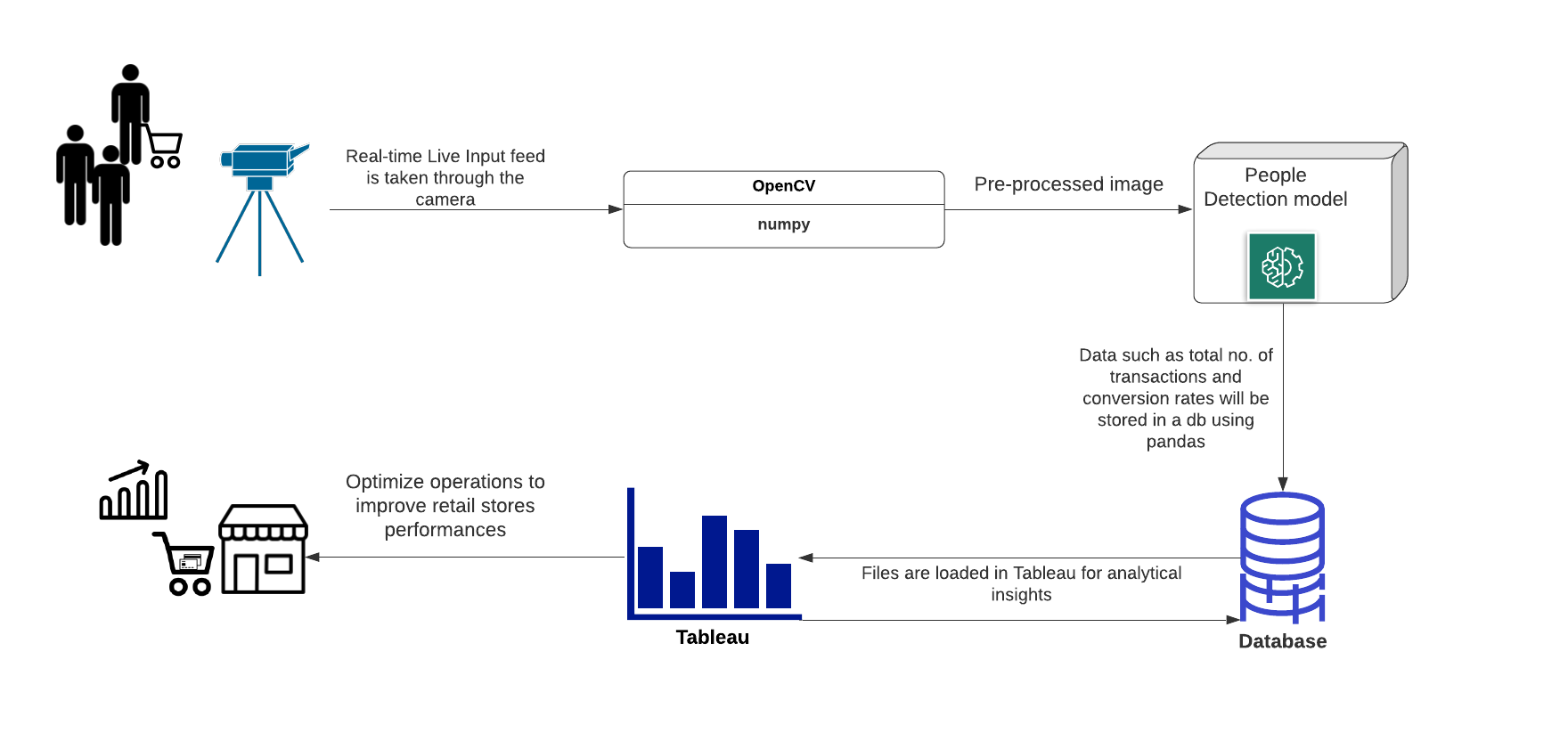}
    \caption{Edge-AI Architecture for people counting system}
    \label{fig:fig1}
\end{figure}
\subsection{Architecture overview}
\begin{itemize}
\item Image capturing and pre-processing steps:
\paragraph{}
We capture the input feed through camera using OpenCV, we are going to resize the image, Convert the image to BGR format which is the format accepted by OpenVINO models and finally reshape the image in [batch size, colour channels, height, width] format with the help of OpenCV and numpy to pass it to OpenVINO model.
bullet.
\item Loading model and counting people:
\paragraph{}
Now as the Preprocessing is done we will load the IR model with OpenVINO and pass the image to it and get detections of humans in the frame along with the count. We will use OpenCV in such a way that the person who has entered the frame and till he leaves the frame, the count for that person has to be one.
bullet.
\end{itemize}
\subsection{Methodology}
In this people counting system {fig2}, we need to either create a human detection model in any AI framework and then convert the model to IR [Intermediate Representation] format with Intel’s distribution of Open VINO Toolkit or use a pre-trained model which is provided by Open VINO. The camera takes the input feed from store’s entrance and send the input feed to the human detection model to perform inference at Edge to count the people. As the number of people coming in = number of people going out so, at the end of the day total number of people counted = number of people counted / 2. We have to enter number of transactions happened in a day to calculate conversion rate.Conversion rate = (total number of transactions / customer traffic) * 100. The data [date,people counted, transactions, conversion rate] will be send to the database everyday automatically. The Data from the database will get exported to Tableau [A Data Visualization tool] for analytical insights and build dashboards to get an overview of how well your store is performing. According to Fashion E-Commerce, Conversion rates will determine the success rate of a store. An online Fashion website has an average of 3 percent of conversion rates whereas, an offline fashion store has an average of 22.5 percent of conversion rates. With these insights on conversion rates, we can improve the store’s operations to perform better using the past generated data for a better future.

\section{People Detection Model}
\begin{figure}[!ht] 
    \centering
    \includegraphics[width=8cm]{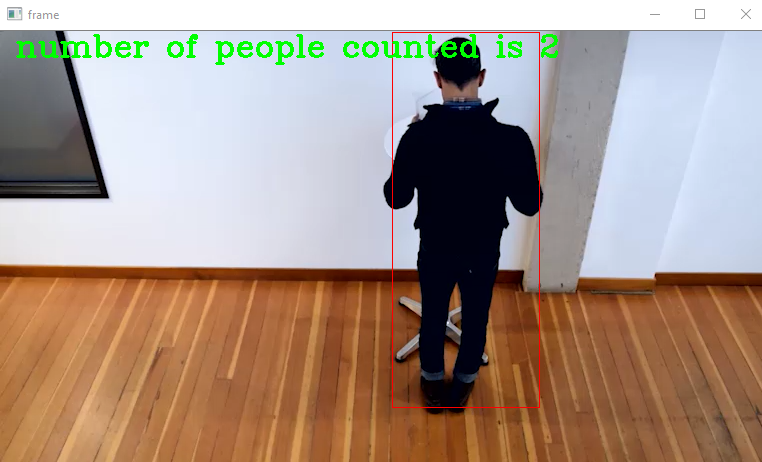}
    \caption{Detecting humans at Edge}
    \label{fig:fig2}
\end{figure}

\ref{fig:fig2}
OpenVINO model zoo has a variety of pre-trained models in Intermediate Representation [IR] format. The IR format of OpenVINO model contains mainly of two files, a ‘.xml file’ and a ‘.bin file’. The .xml file contains the architecture of the model whereas the .bin file contains all the weights and biases of the model.
\paragraph{}
For Human detection, we used person-detection-retail-0013 model from the OpenVINO model zoo, this model is based on MobileNetV2-like backbone which includes depth-wise convolutions because of its faster computational speed compared to regular 3x3 convolutions. This SSD head of 1/16 scale feature map has 12 grouped prior boxes.
\paragraph{}
The person-detection-retail-0013 OpenVINO model has been tested with FP32 precision on Intel’s i3 core processor. The model has successfully detected humans from a video stream given as input.

\section{Case Study}
Among the related research presented in related work section, we did not find scientific contributions focused on the explainability of AI in improving businesses. The architecture we present in this research aims to build a cost-effective people counting system with AI at Edge, where it calculates Conversion rates based on the total number of people counted by the system and number of transactions for the day, which aids in providing analytical insights for fashion store optimization while requiring very minimum hardware at a local level. To build and test the architecture, we worked with real-world data related to the fashion industry, provided by partners from Trendz- a Fashion store in Andhra Pradesh, India. The store has given permission to access CCTV and utilize a system for project
deployment, They have also provided us with the data containing people count, and total transactions for an entire year 2019.
\paragraph{}
We deployed our project in one of the system’s and connected it to the CCTV out front. As the people walked in, the model started to detect humans and counted them. As said, in the methodology, the
number of people walk in = number of people walk out. At the end of the day total number of people counted = number of people counted / 2. We have entered number of transactions happened in a
day to calculate conversion rate. The data [date, people counted, transactions, conversion rate] was send to the database everyday automatically and the trends, conversion rates, peak hours were
displayed in Tableau dashboard{fig3}. The dashboard was updated everyday.
\section{Evaluation}
\begin{figure}[!h] 
    \centering
    \includegraphics[width=12cm]{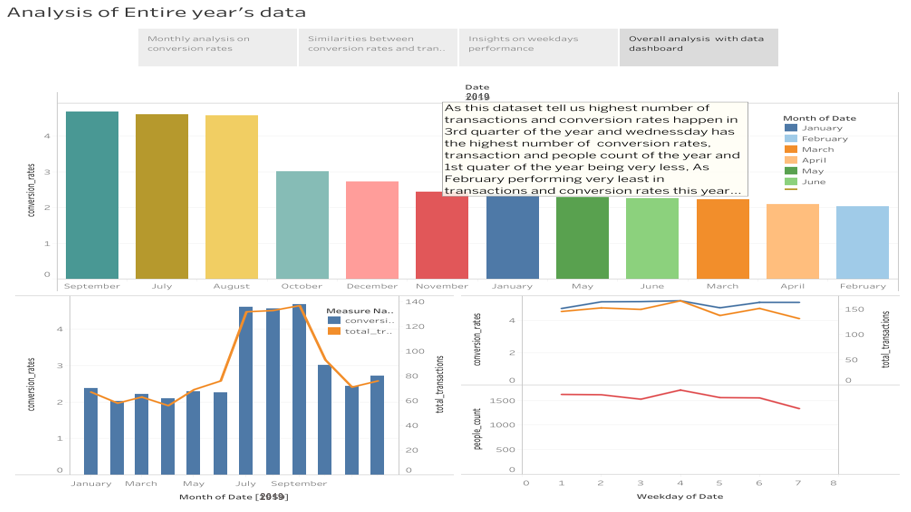}
    \caption{Table dashboard for Trendz.}
    \label{fig:fig3}
\end{figure}

\ref{fig:fig3}
During this period of testing, Store managers were able to identify trends and peak hours and popular products during a particular season. This helped them configure and manage the store accordingly which was proven to very efficient and also brought down the overall operations cost of the store.They were able to manage their stocks  and their marketing accordingly, which brought them a lot of customers.They were able to make much smarter decisions with these insights, when they noticed an increase in sales from July they brought in more staff and attractive stock offers, which gained them maintain this trend till august with high conversion rates.They also took note of the decline in trend during few months and will be changing those collections and bringing out different offers to attract customers even in those months.This can help all the stores optimize their products and operations accordingly achieving better satisfaction to customers as well as profit to the stores.

\section{Conclusion}
A cost effective people counting system which not only counts the number of people but also generates and inputs data in a database with ease. And this is a perfect replacement for all the costly complex sensor technologies and usage of cloud services such as [AWS, Azure, IBM Watson, and Google Cloud Platform] for AI inference. As seen above, that is the basic kind of visualization which provides insights about a particular store, but this is just the tip of the iceberg, we can go way beyond by joining other databases like the products purchased for each transaction for clearer cut insights on what products were being sold mostly. If we are having multiple stores in different locations, by bringing all the databases to one particular place for insights in order to check how well each store is performing, and concentrate on improving the stores which aren’t performing well when compared to others. This will help us in investing on the right products for better sales and most importantly the customer feedback which will help us in improving the store’s performance time to time.

\bibliographystyle{unsrt}

\end{document}